# A Hierarchical Approach to Conditional Random Fields for System Anomaly Detection

Srishti Mishra
PES University

Tvarita Jain
PES University

Dr. Dinkar Sitaram
Professor, PES University

*Abstract*—Anomaly detection or outlier detection to recognize unusual or rare events in large scale systems in a time sensitive manner is critical in many industries, eg. bank fraud, glitches in critical systems, medical alerts, malfunctioning equipment etc. Large-scale systems often grow in size and complexity over time, and anomaly detection algorithms need to adapt to the changing structures. A hierarchical approach can take advantage of the implicit relationships in complex systems and capture anomalies based on context. Furthermore, the features in complex systems may vary drastically in data distribution, capturing different aspects from multiple data sources, and when put together can provide a more complete view of the entire system. Two main datasets are considered, the first consisting of varied system metrics from machines running on a cloud service, and the second of application metrics from a complex distributed software system with inherent hierarchies and interconnections amongst numerous system nodes. Comparing algorithms running in a hierarchical manner, across the Changepoint-based PELT algorithm, cognitive learning-based Hierarchical Temporal Memory algorithms, Support Vector Machines and Conditional Random Fields provides a basis for proposing a Hierarchical Global-Local Conditional Random Field approach to accurately capture anomalies in complex systems, and across various features. Hierarchical algorithms can learn both the intricacies of lower-level or specific features, and utilize these in the global abstracted representation to detect anomalous patterns robustly across multi-source feature data and distributed systems. A graphical network analysis on complex systems can further fine-tune datasets to mine relationships based on available features, which can benefit hierarchical models. Furthermore, hierarchical solutions can adapt well to changes at a localized level, learning on new data and changing environments when parts of a system are over-hauled, and translate these learnings to a global view of the system over time.

## I. INTRODUCTION

Anomalous behavior is inherent to large-scale, enterprise software systems which power a variety of industries from security and IT to energy and healthcare. Anomalies are instances when the behavior of the system is significantly different from the usual and may indicate a problem or unusual activities in the system. In order to predict anomalies, specific patterns of behavior leading up to the anomaly must be identified and used for future prediction. Anomaly detection on real-time streaming data from systems enables corrective action to be taken in critical scenarios thereby saving time, money and personhours.

With the advent of the cloud, these software systems span multiple machines and networks within large-scale data centers, logging large volumes of real-time performance data. The data is often agglomerated from different components of the system at regular intervals taking the form of a time series dataset. The immense amount of data poses a challenge for humans, even experts, to identify anomalies early on.

Turning to machine learning approaches, sequence learning models play an important role in identifying patterns in the dataset which lead to an anomaly. In this paper, hierarchical machine learning approaches are explored to address the large-scale of data originating from multiple sources. An especially interesting hierarchical model, Hierarchical Temporal Memory, lies in the field of cognitive learning algorithms and is compared to hierarchical approaches using traditional machine learning and sequence learning models. Applying hierarchical learning to the problem of large-scale multi-source datasets from enterprise systems, a novel hierarchical approach using a Local-Global Conditional Random Field (CRF) model is proposed as a solution for anomaly detection. Conditional Random Fields are robust for sequence learning and the Local-Global method allows the model to locally learn the idiosyncrasies of each data source as well as globally generalize across the sources and identify anomalies in the system.

The rest of the paper is organized as follows. Section II describes related work in the field of anomaly detection. In Section III, current approaches and models are discussed and compared. Section IV focuses on the proposed approach; detailing the Global-Local CRF model and the motivation behind it. Section V discusses the experimental approach with the nature of the dataset and the evaluation metrics employed. The results of the models are evaluated and compared with the proposed approach, augmented with the network analysis, in Section VI. Section VII concludes with the outcomes of the proposed approach and significant findings from the comparative study.

## II. RELATED WORK

Previous approaches to anomaly detection include both supervised methods, such as support vector machines, regression models, decision trees etc [1,2,3] as well as unsupervised (eg. clustering), however these are yet to be adapted to multi-source, real-time time series datasets. Dimensionality-based methods such as variants of PCA [4,5] are primarily used for high-dimensional, multivariate data streams that can be projected onto a low dimensional space. However, these are restrictive and have strict data constraints, which hinders its adoption in real-world anomaly detection scenarios. Statistical methods such as multivariate statistics [6], Bayesian analysis [7], and frequency and simple significance tests [8] have also been used for anomaly detection. These methods, however, cannot adapt well across multi-source datasets and results get worse as the dataset becomes larger.

Time series analysis such as the ARIMA (Autoregressive integrated moving average) method uses a combination of autoregressive and moving average to model seasonality and predict values [9, 31].

Sequence learning classifiers, such as LSTMs/RNNs models used by Fabian Huch et. all [34] for anomaly detection on imbalanced datasets can identify patterns leading up to an anomaly in the case of predictive anomaly detection. Hierarchical Hidden Markov Models (HHMM) [18] are another sequence prediction model based on nested HMMs to learn a hierarchy of features, where deeply nested markov models learn low-level features and send back predictions and sequences to higher layers, which learn higher-level features. However, as the order of the HMM increases, it gets computationally expensive and is feasible for mostly short-term dependencies.

A few other approaches use statistical methods to extract correlations from the data and events generated from large-scale cloud systems, such as the novel regression-based correlation analysis technique by M. Farshchi et. all [32]. This regression technique uses highly correlated clusters of logs with system metrics to predict expected system values and an observation that significantly deviates from the prediction is classified as an anomaly. D. Sun et. all [33], describe extracting specific features from the system data over a longer detection window and then running a classifier.

In the following sections, key approaches including SVM models, HTM models and changepoint algorithms are discussed with their results, as part of the comparative study with Hierarchical Conditional Random Fields.

### III. CURRENT APPROACHES

#### A. PELT

Another method to detect anomalies is to use changepoint detection algorithms [9] which can identify the occurrence of an anomaly or the start of an anomalous sequence. Due to its reliance on statistical measures of the data, it is computationally expensive to identify changepoints as the time series gets longer. An efficient variant known as PELT changepoint detection [10] has improved the performance of the algorithm, however minimizing false positives remains a challenge. Pruned Exact Linear Time (PELT) is a change point detection algorithm belonging to the time series family. This approach graphically identifies data points that have a significant statistical change and labels them as change points.

Mathematically speaking, consider data points from $z_1, z_2, \ldots, z_n$, if a change point exists at say $z_t$ then there is some statistical change between $\{z_1, \ldots, z_t\}$ and $\{z_{t+1}, \ldots, z_n\}$. The number and position of the points at which the mean changes is inferred. Changepoint detection analyses if the observed results are different and as such it is natural to compare model fits with changepoints to those without. One approach is to use a Likelihood Ratio Test. The likelihood of the model including a change will always provide an improvement over the model with no change, additional parameters improve the fit. If a changepoint is identified, it's position is estimated as

$$t = \text{argmax}\{l(z_1:z_t) + l(z_{t+1}:z_n) - l(z_1:z_n)\}$$

By eye there is often an obvious changepoint at (or by) a time-point s. This means that for any time T ahead of s the most recent change point cannot be at a time t seen before s. This shows that the search step could be pruned and hence avoid searching over any t seen before s. If many t times are pruned, excluded from the minimization then computational time will be drastically reduced and the algorithm becomes very efficient. However the only downside of this change point detection algorithm is that it assumes that any change that is not recent should be pruned. It can be proved that, under certain regularity conditions, the expected computational complexity will be O(n). The most important condition is that the number of changepoints increases linearly with n.

#### B. Hierarchical Temporal Memory

Hierarchical temporal memory (HTM) models are a relatively new development in the field of sequence learning and aim to resemble cortical algorithms found in the human neocortex. They are unsupervised models which can learn from multiple inputs and be trained on streaming data, similar to how the human brain processes information. The HTM algorithm adopts several concepts of learning such as,

a)  Hierarchy of Regions

Neurons are arranged in columnar structures across hierarchy of 6 layers and the dendritic connections between the neurons in each layer ensures that information flows up from the lower layers, composed primarily of sensory inputs and predictions flow downwards from the higher layers to the lower layers.

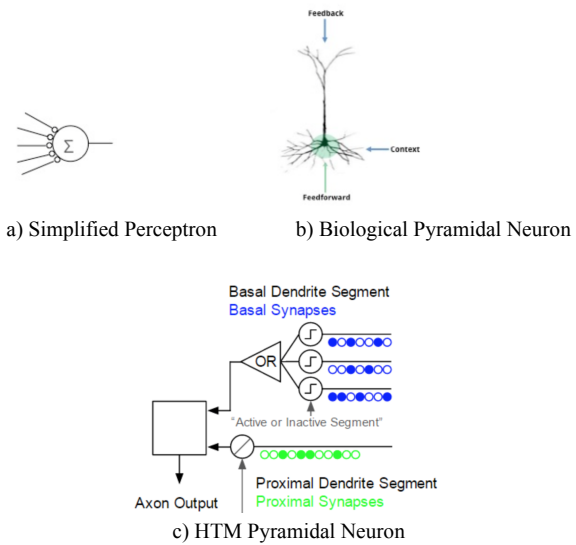

a) Simplified Perceptron   b) Biological Pyramidal Neuron

c) HTM Pyramidal Neuron

Fig. 1.   Comparison of ANN Perceptron, Biological Pyramidal Neuron and HTM Pyramidal Neuron

b)  Pyramidal Neuron Structure

The basic processing unit of the HTM algorithm models a pyramidal neuron [19, 20] which consists of dendritic segments with numerous synapses arranged along dendrites, as seen in Fig. 1 b). Since the perceptron neuron used by ANNs is a simplified model with no dendrites and a few highly precise synapses, (described in Fig. 1 a) and not suitable for temporal sequence learning, other sequence learning models such as RNN/LSTMs [insert reference] too use a special memory cell which is an improvement over the basic perceptron neuron of ANN models.

c)  Sparsity and Input Encoding

Sparsity of neural activity in the human neocortex results in 2% or less neurons being active at any point of time. HTM theory uses sparse representations [23] during encoding inputs as well as in the temporal learning algorithm to capture semantic relationships between inputs.

d) Temporal Learning

HTM models learn and recall high-order sequences in an unsupervised manner using online learning. The Spatial Pooling algorithm uses synaptic learning and Hebbian learning to convert the SDR inputs to a normalized pattern of active columns [22]. Temporal context is captured by transitions between cells or columns within the layer and from lower layers using the pyramidal neuron structure.

e) Anomaly Classifier

As HTM models do not explicitly model anomalous data points, an additional classification layer is used to compute the likelihood of an anomaly by comparing the sparse vector prediction with the observed value in real-time [12].

C. Support Vector Machines

A variant of SVMs known as One-Class SVMs [14] can handle imbalanced datasets and learn in an unsupervised manner which is generally the case in real-world anomaly detection datasets. In contrast to traditional SVMs, One-Class SVMs learn a decision boundary that achieves maximum separation between the samples of the known class and the origin. Only a small fraction of data points are allowed to lie on the other side of the decision boundary: those data points are considered as outliers.

The reason for choosing OCSVM for anomaly detection was that in a typical two class SVM, it was not possible to represent all the faulty states of any system and moreover to simulate a faulty environment was nearly impossible or quite expensive.

In the case of OCSVM, by just providing the normal training data, the algorithm creates a (representational) model of this data. If newly encountered data is too different, according to some measurement, from this model, it is labeled as out-of-class. The previous HTM approach and SVM approach both use training data to learn the normal class and any point classified outside this class is considered an anomaly or an outlier. The nu and gamma parameters in the Radial basis function (RBF) kernel of the OCSVM help limit misclassifications during training and control overfitting respectively. One class SVMs were built upon the original Support Vector Method For Novelty Detection by Schölkopf et al. [1] which essentially separates all the data points from the origin (in feature space F) and maximizes the distance from this hyperplane to the origin. This results in a binary function which captures regions in the input space where the probability density of the data lives. Thus the function returns +1 in a "small" region (capturing the training data points) and −1 elsewhere.

IV. PROPOSED APPROACH

A. Motivation

Identifying anomalies early on in the time series data collected from enterprise systems overlaps with the domain of sequence learning problems. Real-time performance data contains hidden patterns across varying length scales which lead up to or possibly cause an anomaly. The time series data comprises multiple features or performance metrics collected from numerous related data sources in large-scale systems.

Consider a time series dataset consisting of a sequence of observations and class labels represented by $X = \{(x_k, y_k) | k\}$, with multiple features where $X \in D = \{d_1, d_2 \ldots d_j | j\}$. In order to predict the current class label at observation $x_t$, sequence learning models learn relationships between observations and classes at previous timesteps such as $x_{t-1}$ in 1-gram models or $x_{t-2}$, $x_{t-1}$ for 2-gram models and so on.

Generative models such as HMMs and Naive Bayes are a class of sequence learning models which model the joint probability distribution of the inputs and classes to generate sequences. However, these generative models often fail to model overlapping features due to the Naive Bayes independence assumption and become computationally infeasible as the number of observations increases.

Conditional Random Fields (CRFs) are discriminative models which model the conditional probability to find the boundary between classes, during classification. When investigating anomalies, the cause of the anomaly is often implicitly present in various performance metrics collected from the system. CRFs model relationships between these overlapping and non-independent features **D**, as seen in the Figure 2, using factors to ensure robust prediction of the sequence of labels based on the input sequences. Furthermore, to reduce the rate of false alarms, the confidence measure along with every class label prediction can be used to filter out low confidence predictions.

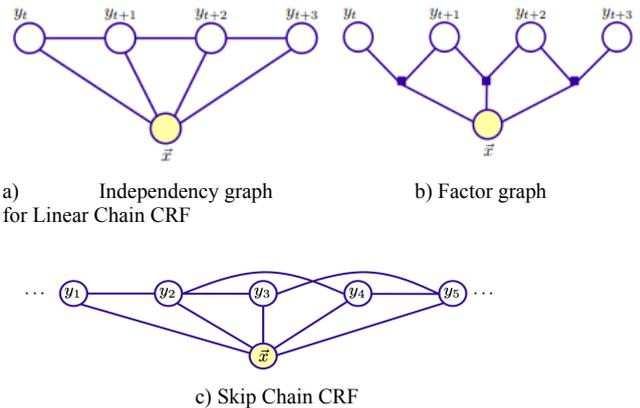

a) Independency graph for Linear Chain CRF
b) Factor graph

c) Skip Chain CRF

Fig. 2. Linear Chain Conditional Random Field describing the probability distribution [27]

The factors in the graph correspond to feature functions represented by ψ and the probability distribution of the output sequence is modeled as follows,

$$p(\underline{y} | \underline{x}) = \frac{1}{Z(\underline{x})} \prod_{j=1}^{N} \psi_j(\underline{x}, \underline{y})$$

The feature function in the CRF is a log linear based function to model the conditional probability. It models the probability of seeing a particular class label given the current input along with the likeliness of the current sequence of labels based on previous labels. The feature set provided to the function is made richer by including features of previous observations from the set **D**. The probability distribution proves expensive to compute due to the normalization constant **Z**; all possible values of the feature functions. CRFs model temporal context through arbitrary structures, seen in the skip chain CRF in Figure 2 c), however, the Global-Local CRF approach (Fig 3) uses the

Linear Chain CRF (Figure 2 a) ) which is computationally feasible. The Local Linear Chain CRF considers local dependencies between adjacent segments as follows,

$$p(\underline{y}|\underline{x}) = \frac{1}{Z(\underline{x})} exp\left( \sum_{k=1}^{K} \lambda_i f_i \left( x_{k-1}, x_k \right)_{y_k} + \sum_{k=1}^{K} V_{y_k} \right)$$

The presence of multiple sources of related data introduces a natural hierarchy into the dataset, where relationships between different data sources can be modeled as well as relationships between features within a data source. The Global CRF models a feature rich set of observations including both original features as well as indicative features generated by local CRF models. In the hierarchical Global-Local CRF approach, the Local CRF models can be compared to lower-level regions which learn localized representations and data-source specific anomalies, while the Global CRF model acts as a higher-level region which learns an abstract representation of anomalies across all sources, described in Figure 3.

The computational feasibility of running a Global-Local CRF model on large-scale software systems is aided by the hierarchical structure of the model. When data is collected over long periods of time or as the number of systems increases, a hierarchical data center offers a scalable solution to parallelize computation among the localized models. Moreover, a hierarchical approach can learn to catch anomalies across the system at a macro-scale, even if the amount of training data varies from local micro-scale components.

### B. Description of Model

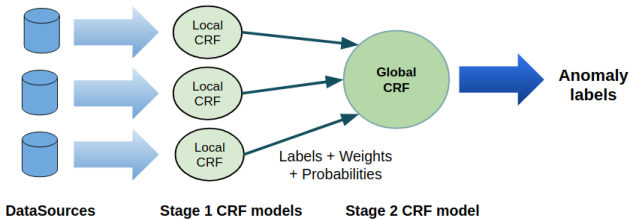

Fig. 3. Global-Local CRF approach

*Overview*
The Global - Local CRF model works in two stages. In the first part anomalous data points are classified within each data source. These outputs are used in the second stage to determine anomalies across the data sources.

*Early and Late Fusion*
Neural network fusion refers to using a combination of models and is suitable for modeling multi-source datasets. In early fusion, all features from all the data sources are consolidated and fed to a single global model, helping capture hidden relationships between sources. However, in large-scale systems the multiple sources of data could be independent of each other or exhibit misleading correlations and an early fusion model may not find meaningful relationships between the features. Furthermore, the immense amount of data generated in large-scale systems from each source further increases the computational complexity of running a single model and excessive unrelated features may be detrimental to the results.

In a late fusion approach, a separate model runs on each source of data and the outputs received from these individual sources are fed as features to a global model which computes results across these multiple sources. This could increase the complexity overall as the number of models increases, but pruning and collating sources based on relationships and patterns in data can help keep the number of models at an optimal number.

*Joint and Separate Training*
Assessing computationally, a hierarchical model can undergo either joint training or separate training. In joint training, all models from each source are together in a streaming manner. In the Global-Local CRF, a single iteration would train both the local models and the global model, with the stream of predictive outputs from the Local models being continuously fed to the Global model. Separate training is akin to batch training a subset of models in a single iteration and the consolidated batch output is used separately by the Global model. In the scenario that each iteration consists of a single observation, separate training behaves similar to joint training.

TABLE I. SUMMARIZED FEATURES

|  | Features |
|---|---|
| Dataset 1: AWS Server Metrics | **(System)** <br> - CPU utilization <br> - Incoming network traffic (bytes) <br> - Outgoing network traffic (bytes) |
| Dataset 2: Enterprise Software Application Metrics | **(Database)** <br> - Currently active connections <br> - Database connection activity <br> - Database connection delay <br> - Failed reserve connection requests <br> - Relative unavailable connections <br> - Reserve connection request activity <br> - Started/Failed/Successful connections <br> - Currently active transactions <br> - Transaction rollback/commit activity <br> **(System)** <br> - Activity in JVM memory spaces <br> - Physical memory activity <br> - Relative JVM memory space usages <br> - Relative physical memory usage <br> - Daemon/Total thread count <br> - Stuck threads <br> - Relative swap usage <br> - Swap activity <br> - Class loading/unloading activity <br> - CPU usage/time <br> - Garbage collection activity/duration <br> - Hit rate of prepared statement caches <br> - JVM objects to be finalized <br> - Missing data <br> - Relative open file descriptors <br> - Relative physical memory usage |

## V. EXPERIMENTS

### A. Description of Datasets

The anomaly detection models are tested on two types of data from enterprise software systems. The first dataset comes from Numenta Anomaly Benchmark (NAB) [12] data corpus consisting of Amazon Web Services (AWS) server metrics from related sources in an EC2 server. The data was collected using AWS CloudWatch service which monitored

a) CPU utilization of the system
b) Number of bytes of incoming network traffic to the system and,
c) Number of bytes written to the disk in the system.

TABLE II. DATA IMBALANCE

|  | % of Anomalies | Train/Test Split Ratio |
|---|---|---|
| Dataset 1: AWS Server Metrics | 3.2% | .45 : .55 |
| Dataset 2: Enterprise Software Application Metrics | 2.6 % | .45 : .55 |

As the dataset was imbalanced with only about 3% anomalous instances, outlier data points were labeled as anomalies using a strict variant of the IQR outlier-detection method. Additionally, the train-test split was kept at a ratio of .45: .55 due to the sparsity of anomalies throughout the dataset. Table II indicates the class distribution of anomalous data points to normal data points for both datasets.

The second dataset is obtained from an enterprise system software [27] and consists of multi-source, multi-node data from an enterprise software application running on a distributed network of systems. The data was collected from the operating system and WebLogic Server monitoring beans. The distributed enterprise system was formed by 10 nodes running 20 instances of the application overall. There were 253 metrics covering CPU utilization, memory usage, swap activity from the operating system, as well as database activity, transactions, JVM object memory etc from the software application as summarized [26]. The 253 features were ranked in order of importance using a Gradient Boosted Machine model [28] and a subset of features were selected for each data source. Furthermore, to handle the imbalance, certain time windows of data containing an adequate number of anomalies were selected for initial experiments.

The metrics from both datasets are summarized in Table I. In the first dataset, the Global-Local CRF model exploits the multi-source aspect of the data by running local models on each source, i.e, CPU utilization, Network traffic and Disk activity separately, and then using a global model to identify system-wide anomalies. The second dataset contains multi-source data from around ten distributed systems. In this case, local models identify anomalies on each node in the distributed system and a global model identifies anomalies across the distributed system, which may indicate critical system-failure. A graphical analysis of the connections between nodes in the distributed system highlights clusters of nodes that are closely related and may fail together. The hierarchical and modular nature of the Global-Local CRF ensures it runs first on each node and then on each cluster of nodes to identify local anomalies and then potential failures across the system.

B. Evaluation Metrics

1. Cohen kappa score :

The Cohen-kappa statistic is a great way to handle both multi-class and imbalanced class problems which is prominent in both the datasets used here.
Mathematically speaking,
Cohen Kappa is defined as :

$$\kappa = \frac{p_o - p_e}{1 - p_e} = 1 - \frac{1 - p_o}{1 - p_e},$$

Here, Po is the Observed Result whereas Pe is the expected result. This statistic is a quantitative measure of reliability of how a given classifier performs compared to another classifier, corrected for how often that the two may agree by chance. Practically, it captures how the classifier of interest performs over a classifier that would guess randomly based on the frequency of the classes.

2. Confusion matrix, Accuracy, Detection Rate:

Additionally, a confusion matrix was used to measure the model's performance, and summarize the prediction results of the classification problem. The detection rate identifies the proportion of points identified as anomalies that are true anomalies in the system, very useful at a local level. To keep false detections low, at least on a global scale, the False Positive rate is also measured, i.e., the proportion of points that are not anomalies but are identified as anomalies.

TABLE III. COMPARISON OF RESULTS (DATASET 1)

| Models | Accuracy Measures | | | | |
|---|---|---|---|---|---|
|  | Model name | Training Accuracy[a] | Testing Accuracy[a] | Detection Rate | False Positive Rate |
| Global-Local CRF Model | Local CRF Model A (CPU Utilization) | 98.9 % | 100.0 % | 100% | 0% |
|  | Local CRF Model B (Disk Writes in Bytes) | 96.9 % | 89.8 % | 53.1 % | 0.38 % |
|  | Local CRF Model C (Incoming Network Traffic in Bytes) | 39.9 % | NA[b] | NA[b] | NA[b] |
|  | **Global CRF Model (Late Fusion)** | 95.8 % | 89.0 % | 81.7 % | .05 % |
| Local HTM Models | Local HTM Model A (CPU Utilization) | NA[c] | 13.3% | 50.0 % | 0.33 % |
| | Local HTM Model B (Disk Writes in Bytes) | NA[c] | 6.08 % | 4.83 % | 0.59 % |
| | Local HTM Model C (Incoming Network Traffic in Bytes) | NA[c] | NA[b] | NA[b] | NA[b] |
| Local PELT Changepoint Models | Local PELT Model A (CPU Utilization) | 0.0 % | 0.01 % | 0.62 % | 14.15 % |
| | Local PELT Model B (Disk Writes in Bytes) | 2.93 % | 0.62 % | 16.1 % | 20.1 % |
| | Local PELT Model C (Incoming Network Traffic in Bytes) | 0.0 % | NA[b] | NA[b] | NA[b] |
| Local SVM Models (rbf kernel) | Local SVM Model A (CPU Utilization) | 19.3 % | 66.7 % | 96.4 % | 0 |
| | Local SVM Model B (Disk Writes in Bytes) | 44.5 % | 47.2 % | 95.2 % | 1.96 % |
| | Local SVM Model C (Incoming Network Traffic in Bytes) | 0.505% | NA[b] | NA[b] | NA[b] |

[a] Cohen Kappa score is used as Accuracy
[b] Testing sample contained no anomalies
[c] HTM model uses online learning

## VI. RESULTS

To analyze the performance of the Global-Local CRF Model, the first dataset consisting of system metrics is compared with existing techniques including anomaly detection sequence-based models, changepoint models and an SVM classification model. The proposed CRF approach is further tested on the 2nd dataset of enterprise software system metrics with complex hierarchical relationships.

### A. Dataset 1

The results from the single node, multi-source data set with a single feature is a special simplified case of the Hierarchical Global-Local CRF model and are comparable to the localized models of HTM and PELT models. Note that HTM models use online learning and hence do label any training data. The PELT changepoint mode did not perform well on any of the metrics data, possibly due to the variable nature of the data. The assumption that a marked (sudden increase/decrease) change of the metrics time series in a system is an anomaly is false in this case. In this case, the sequence or fluctuations of metrics data leading up to the anomaly provides more insight. The HTM models, which assume that there are distinctive patterns leading up to an anomaly, classify each point into 3 classes: High, Medium and Low with a corresponding score. To convert this to a binary classification, an additional classification layer used the continuous-scale anomaly score to determine if a point is an anomaly or not.

### B. Dataset 2

The multi-source dataset, collected from 10 system nodes each running 2 instances of the application, has mixed results. The results from the multi-node, multi-source data set with a single feature is a special simplified case of the Hierarchical Global-Local CRF model and are comparable to the localized models of HTM and PELT models. Note that HTM models use online learning and hence do label any training data. Running all the algorithms in a late fusion model, first working at a system-level and then with combined cues, led to poor results, and the Global Local CRF model performed at a less than 50% accuracy as well. This indicated that there might be further underlying relationships at a system level that are not captured by the Linear Chain CRF. It could indicate that each node may not be completely independent of the other. Analysis of patterns hidden in the system metrics may provide a better understanding of the hierarchical nature of the complex enterprise system.

### Network Analysis

In a distributed enterprise system, applying hierarchical learning on existing dependencies on clusters of nodes can help monitor the global system more accurately. To mine relationships between the system nodes in the enterprise system, features containing database connection information helped uncover indirect dependencies and relationships.

The multi-node system in question consists of host nodes and several database sources. Clusters of nodes show strong relationships in terms of either active connections to a single source, or errored unavailable connections to a source.

This provides insights into the causation of the anomaly as well as a blueprint for hierarchical models to follow. In the Hierarchical Global-Local CRF for this complex system, the predictions and features from single nodes in Stage 1 are fed into multiple Global models in Stage 2 consisting of clusters of closely related nodes. In this case, a 2-level hierarchy proves to run efficiently and catch anomalies, but in more complex systems, multi-level hierarchies could influence the model, and further thresholding can be applied based on the heuristics or logical view of complex systems.

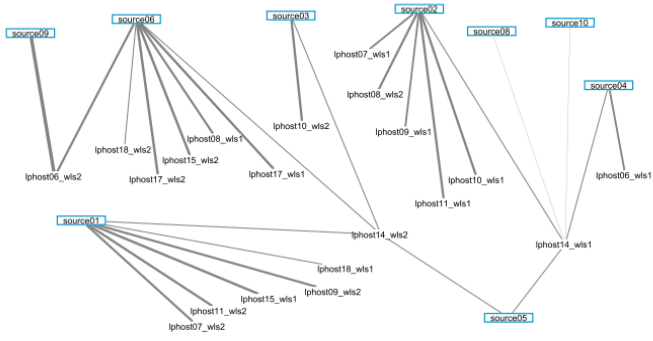

a) Dependencies between hosts based on Unavailable Connections from Source databases

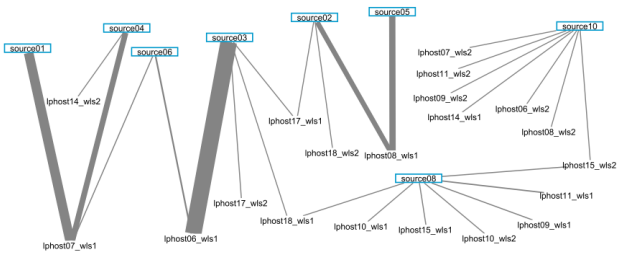

b) Dependencies between hosts based on Active Connections from Source databases

Fig. 4. Dependency Graphs of Multi-Node systems based on Source Databases

*Network-based Results*

Combining nodes associated with Sources 02, 03, 04, and 05, and considering those with enough clean data, these single nodes can be considered as local CRF models. The global CRF model consists of 4 clusters, as per the results in Table IV. An anomaly in any of these 4 can be considered detrimental to the system, as it affects more than 1 node. The accuracy on average was above 90% per cluster, a major improvement from less than 50% accuracy for single nodes, indicating that without hierarchical models and understanding the implicit underlying relationships, it would be difficult to form an accurate predictive model of the complex system from a local view alone.

## VII. CONCLUSIONS & FUTURE WORK

Hierarchical learning approaches attempt to mirror the intrinsic relationships in many systems, as well as learn from localized models allowing for a more comprehensive prediction mechanism. To capture the effect of localized nodes or clusters of nodes on an entire system, hierarchy-based models can benefit by taking a local-global approach and work well across a diverse feature set. The limitations of 2-level hierarchy could be seen in the complex system dataset in the Hierarchical CRF approach. For simpler systems with fewer features, a single global system anomaly could be detected clearly. However, detecting anomalies in multiple clusters at the $2^{nd}$ level as system complexity increases provides an opportunity to model multi-level hierarchies, or apply thresholding across multiple clusters at the $2^{nd}$ level to capture the global system state better. Conditional Random Fields have proven to work as well as or better than other models, including cognitive-based models, time series and SVM models. Moreover, the dual CRF manages to capture the implicit relationships in both localized temporal data as well as in the larger hierarchical systems. Only linear CRFs were considered here, however, fine-tuning the structure of the CRF for more complex systems could lead to better results as well.

TABLE IV. HIERARCHICAL GLOBAL – LOCAL CRF MODEL (DATASET 2)

| Models | Accuracy Measures | | | | |
|---|---|---|---|---|---|
| | Sources | Training Accuracy[a] | Testing Accuracy[a] | Detection Rate | False Positive Rate |
| Global-Local CRF Model | Source03<br>lphost10_wls2, lphost14_wls2 | 98.0% | 92.5% | 95.5% | 2.79% |
| | Source04<br>lphost11_wls2, lphost15_wls1,<br>lphost18_wls1, lphost14_wls2 | 96.0% | 90.8% | 89.0% | 0.80% |
| | Source02<br>lphost09_wls1, lphost11_wls1,<br>lphost10_wls1, lphost14_wls1 | 80.8% | 88.0% | 94.6% | 4.9% |
| | Source05<br>lphost14_wls2, lphost14_wls1 | 99.4% | 99.0% | 100% | 0.51% |

## REFERENCES

1. Bernhard Schölkopf, Robert Williamson, Alex Smola, John Shawe-Taylor, and John Platt. 1999. Support vector method for novelty detection. In Proceedings of the 12th International Conference on Neural Information Processing Systems (NIPS'99). MIT Press, Cambridge, MA, USA, 582–588..

3. Chandola, Varun, Arindam Banerjee, and Vipin Kumar. "Anomaly detection: A survey." ACM computing surveys (CSUR) 41.3 (2009): 15.
4. V. Chandola, V. Mithal, V. Kumar. Comparative evaluation of anomaly detection techniques for sequence data. Proceedings of the 2008 Eighth IEEE International Conference on Data Mining (2008), pp. 743-748.
5. M.A.F. Pimentel, D.A. Clifton, L. Clifton, L. Tarassenko. A review of novelty detection. Signal Process., 99 (2014), pp. 215-249,
6. Lee Y.J., Y.R. Yeh, Wang Y.C.F. Anomaly detection via online oversampling principal component analysis. IEEE Trans. Knowl. Data Eng, 25 (2013), pp. 1460-1470.
7. A. Lakhina, M. Crovella, C. Diot. Diagnosing network-wide traffic anomalies. ACM SIGCOMM Comput. Commun. Rev, 34 (2004), p. 219.
8. Taylor, Carol, and Jim Alves-Foss. Low cost network intrusion detection. (2000).
9. Barbara, Daniel, Ningning Wu, and Sushil Jajodia. Detecting Novel Network Intrusions Using Bayes Estimators. SDM. 2001.
10. A.M. Bianco, M. García Ben, E.J. Martínez, V.J. Yohai, Outlier detection in regression models with ARIMA errors using robust estimates, J. Forecast. 20 (2001) 565–579.
11. M. Basseville, I. V Nikiforov, Detection of Abrupt Changes, 1993.
12. Gachomo Dorcas Wambui, Gichuhi Anthony Waititu, Anthony Wanjoya. The Power of the Pruned Exact Linear Time(PELT) Test in Multiple Changepoint Detection.American Journal of Theoretical and Applied Statistics.Vol.4, No. 6, 2015, pp. 581-586.
13. Qin, Min, and Kai Hwang. Frequent episode rules for intrusive anomaly detection with internet datamining. USENIX Security Symposium. 2004.
14. Ahmad, S., Lavin, A., Purdy, S., & Agha, Z. Unsupervised real-time anomaly detection for streaming data. Neurocomputing 262 (2017): 134-147.
15. Cui, Yuwei, Subutai Ahmad, and Jeff Hawkins. "Continuous online sequence learning with an unsupervised neural network model." Neural computation 28.11 (2016): 2474-2504.
16. Zhang, Ming, Boyi Xu, and Jie Gong. "An anomaly detection model based on one-class SVM to detect network intrusions." 2015 11th International Conference on Mobile Ad-hoc and Sensor Networks (MSN). IEEE, 2015.
17. Malhotra, Pankaj, et al. "Long short term memory networks for anomaly detection in time series." Proceedings. Presses universitaires de Louvain, 2015.
18. Song, Yale, et al. "One-class conditional random fields for sequential anomaly detection." Twenty-Third International Joint Conference on Artificial Intelligence. 2013.
19. Fine, Shai, Yoram Singer, and Naftali Tishby. "The hierarchical hidden Markov model: Analysis and applications." Machine learning 32.1 (1998): 41-62.
20. Huang, Qixing, et al. "A hierarchical conditional random field model for labeling and segmenting images of street scenes." CVPR 2011. IEEE, 2011.
21. Poirazi, Panayiota, Terrence Brannon, and Bartlett W. Mel. "Pyramidal neuron as two-layer neural network." Neuron 37.6 (2003): 989-999.
22. J. Hawkins and S. Ahmad, "Why neurons have thousands of synapses, a theory of sequence memory in neocortex," Front. Neural Circuits, vol. 10, Mar. 2016.
23. Cui, Yuwei, et al. "A comparative study of HTM and other neural network models for online sequence learning with streaming data." 2016 International Joint Conference on Neural Networks (IJCNN). IEEE, 2016.
24. J. Hawkins, S. Ahmad, and D. Dubinsky, "Cortical learning algorithm and hierarchical temporal memory," Numenta Whitepaper, 2011. [Online].Available:http://numenta.org/resources/HTM_CorticalLearningAlgorithms.pdf
25. Ahmad, Subutai, and Jeff Hawkins. "Properties of sparse distributed representations and their application to hierarchical temporal memory." arXiv preprint arXiv:1503.07469, 2015.
26. Kanerva, Pentti. Sparse distributed memory. MIT press, 1988.
27. Numenta Anomaly Benchmark. [Online]. Available: https://github.com/numenta/NAB
28. F. Huch. Repository for feature data. [Online]. Available: https://www.kaggle.com/anomalydetectionml/features
29. Klinger, Roman, and Katrin Tomanek. Classical probabilistic models and conditional random fields. TU, Algorithm Engineering, 2007.
30. Xu, Zhixiang, et al. "Gradient boosted feature selection." Proceedings of the 20th ACM SIGKDD international conference on Knowledge discovery and data mining. ACM, 2014.
31. Sutton, Charles, and Andrew McCallum. "An introduction to conditional random fields." Foundations and Trends® in Machine Learning 4.4 (2012): 267-373.
32. H. Zare Moayedi and M. A. Masnadi-Shirazi, "Arima model for network traffic prediction and anomaly detection," 2008 International Symposium on Information Technology, Kuala Lumpur, 2008, pp. 1-6
33. M. Farshchi, J. Schneider, I. Weber and J. Grundy, "Experience report: Anomaly detection of cloud application operations using log and cloud metric correlation analysis," 2015 IEEE 26th International Symposium on Software Reliability Engineering (ISSRE), Gaithersbury, MD, 2015, pp. 24-34.
34. D. Sun, M. Fu, L. Zhu, G. Li and Q. Lu, "Non-Intrusive Anomaly Detection With Streaming Performance Metrics and Logs for DevOps in Public Clouds: A Case Study in AWS," in IEEE Transactions on Emerging Topics in Computing, vol. 4, no. 2, pp. 278-289, April-June 2016.
35. F. Huch, M. Golagha, A. Petrovska and A. Krauss, "Machine learning-based run-time anomaly detection in software systems: An industrial evaluation," 2018 IEEE Workshop on Machine Learning Techniques for Software Quality Evaluation (MaLTeSQuE), 2018, pp. 13-18, doi: 10.1109/MALTESQUE.2018.8368453.